\documentclass{article}
\pdfoutput=1                

\usepackage{arxiv}          

\usepackage[utf8]{inputenc} 
\usepackage[T1]{fontenc}    
\usepackage{hyperref}       
\usepackage{url}            
\usepackage{booktabs}       
\usepackage{amsfonts}       
\usepackage{nicefrac}       
\usepackage{microtype}      

\usepackage{float}          
\usepackage{gensymb}        
\usepackage{svg}            
\usepackage{amssymb}        
\title{A Mask R-CNN approach to counting bacterial colony forming units in pharmaceutical development} 

\author{Tanguy Naets$^1$, Maarten Huijsmans$^1$, Paul Smyth$^2$, Laurent Sorber$^1$,  Ga\"el de Lannoy$^2$
\vspace{.3cm}\\
1- Radix.ai, Brussels, Belgium
\vspace{.1cm}\\
2- GSK Vaccines, Rixensart, Belgium\\
}

\begin{document}
\maketitle

\begin{abstract}

We present an application of the well-known Mask R-CNN approach to the counting of different types of bacterial colony forming units that were cultured in Petri dishes. Our model was made available to lab technicians in a modern SPA (Single-Page Application). Users can upload images of dishes, after which the Mask~R-CNN model that was trained and tuned specifically for this task detects the number of BVG- and BVG+ colonies and displays these in an interactive interface for the user to verify. Users can then check the model's predictions, correct them if deemed necessary, and finally validate them. Our adapted Mask~R-CNN model achieves a mean average precision (mAP) of 94\% at an intersection-over-union (IoU) threshold of 50\%. With these encouraging results, we see opportunities to bring the benefits of improved accuracy and time saved to related problems, such as generalising to other bacteria types and viral foci counting.
\end{abstract}


\section{Introduction}
\label{Introduction}

In vaccine development, properly counting and differentiating between different types of bacteria colonies is a key step. It is a time consuming and error-prone task that prevents biologists from performing more meaningful and fulfilling work. Several approaches have been investigated to try to reduce the biologists' burden of manually counting such colonies. In the field of traditional computer vision, several authors \cite{barber2001automated, niyazi2007counting, brugger2012automated, torelli2018autocellseg} developed algorithms that try to threshold Petri dish images into binary masks of background and CFUs (Colony-Forming-Units), followed by various post-processing techniques to try to segment each CFU. Open-source-software such as OpenCFU \cite{geissmann2013opencfu}, CellProfiler \cite{mcquin2018cellprofiler} and AutoCellSeg \cite{torelli2018autocellseg} arose from research in this field, while commercial software and hardware solutions also exist. These approaches can suffer from some drawbacks:

\begin{itemize}
    \item a lack of multi-class CFU counting
    \item poor performance on overlapping CFUs and miscounting of agglomerated colonies \footnote{\cite{boukouvalas2019automatic} provides encouraging results using cross-correlation granulometry to try to estimate the quantity of colonies in agglomerates.}
    \item the need for careful calibration and good exposure conditions to properly exclude the background.
\end{itemize}

In computer vision, deep learning algorithms can roughly be classified in four categories depending on the type of task they seek to solve:
\begin{itemize}
    \item \emph{image classification/regression}: predicts one or several variables\footnote{Variables can be nominal, ordinal, and/or quantitative.} for an entire given image
    \item \emph{image segmentation/semantic segmentation}: predicts one or several variables for each pixel of a given image
    \item \emph{object detection}: detects objects' bounding boxes in a given image, and can further classify them
    \item \emph{instance segmentation}: is a hybrid of the latter two categories, which detects and classifies objects in an image, as an object detection algorithm would do, and further binarizes the pixels within each bounding box into object and background pixels, in a similar fashion to image segmentation.
\end{itemize}

In recent years, deep learning approaches for CFU counting and viral foci counting, as well as in many other fields of the biomedical sector, have became increasing common, with the U-Net \cite{ronneberger2015u} architecture being particularly popular. To our knowledge \cite{ferrari2017bacterial} was one of the first papers to make use of a classical convolutional neural network instance segmentation algorithm for bacterial colony counting. In 2018-2019 Beznik et al., including a subset of our authors, developed the application of U-Net based models to count colonies of two different bacteria classes \cite{Beznik2019,Beznik2020}. By 2019, \cite{lin2019counting} was tackling CFUs counting with a U-Net architecture and hosting their model in the cloud to make it directly available for users taking pictures with mobile phones. 

The are two main challenges faced when counting colonies using image segmentation algorithms: 

\begin{itemize}
    \item \emph{strongly imbalanced classes}: most pixels are background, requiring carefully constructed loss functions in order to obtain satisfactory predictions of colony pixels.
    \item \emph{overlapping CFUs}: separating CFUs during post-processing of the colonies masks is often achieved based on same-class-pixels' connectivity. Therefore, as soon as two colonies of the same class are connected by one or more pixels, these CFUs are misleadingly considered as a single colony.
\end{itemize}

Inspired by UNet \cite{ronneberger2015u} and the 2018 Kaggle Data Science bowl \cite{Caicedo2019}, Beznik et al. \cite{Beznik2019} attempt to address overlapping CFUs by labelling images with an additional boundary class. This breaks the connectivity of overlapping colonies and agglomerates of CFUs can then be separated with processing techniques. Unfortunately this technique increases the data imbalance, as the newly introduced class of boundary pixels is even less common than the CFU classes and as soon as boundaries are not predicted perfectly, overlapping colonies are reconnected and one can no longer distinguish the single entities of an agglomerate.

This paper addresses the two challenges described above by tackling the problem with an \emph{instance segmentation} architecture. At the time of starting this work, the latest major breakthrough in the field of instance segmentation occurred in 2017 with the publication of Mask~R-CNN \cite{he2017mask}. Mask~R-CNN is an instance segmentation algorithm, i.e., a combination of object detection and semantic segmentation algorithms. We believe that this approach should be immune to imbalanced classes and overlapping CFUs, as it need not classify background pixels, nor does it need to rely on a boundary class to make the distinction between overlapping objects. Furthermore, by focusing on instance segmentation to generate a mask of each CFU, in addition to bounding boxes, we could later estimate the quantity of bacteria in each of those CFUs.

In this work, we develop our model using the robust and efficient Matterport implementation of Mask~R-CNN \cite{matterport_maskrcnn_2017} and train on 101 GSK laboratory images of Petri dishes containing two types of bacterial colonies, gathered and provided to us by \cite{Beznik2019}. In addition to building a efficient CFU differentiator and counter, we built a modern web application so that our model can be effectively used in production by lab technicians to automate the counting, as well as related downstream tasks.

\section{Model development}
\label{deep-learning}

\begin{figure}
    \centering
    \includegraphics[width=0.9\textwidth]{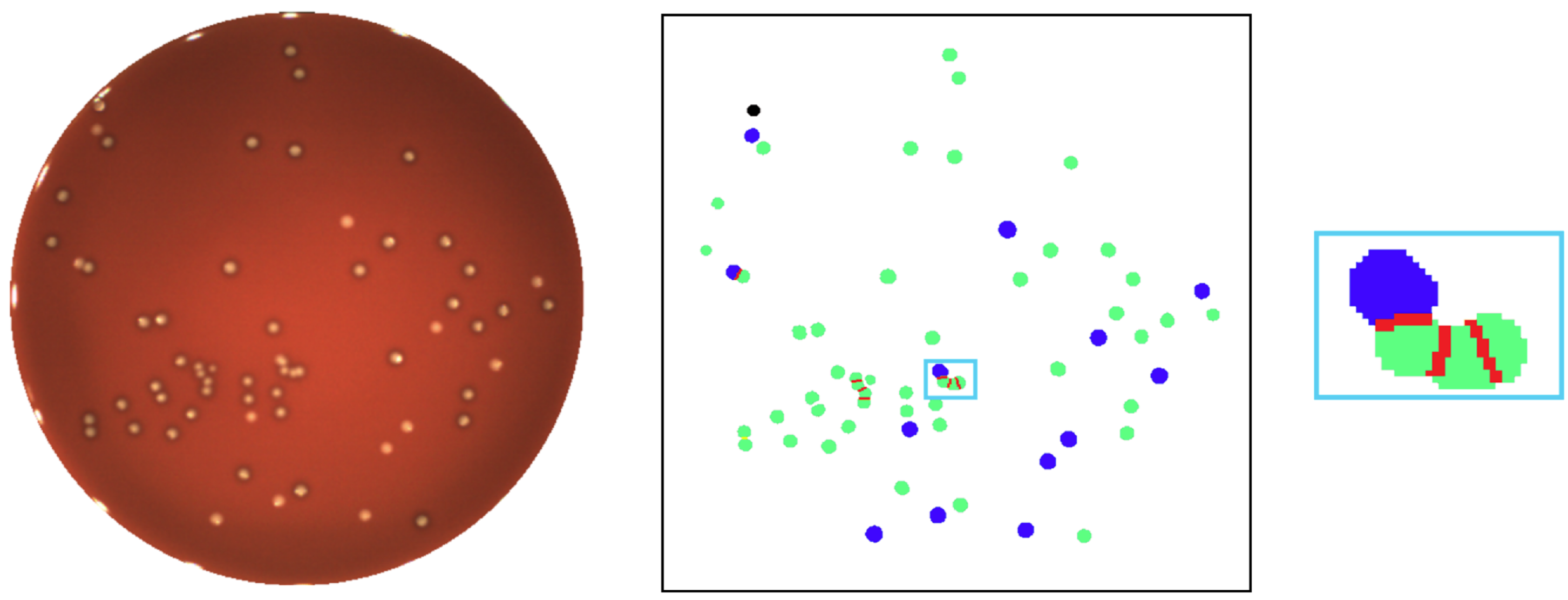}
    \caption{From left to right, an image of a Petri dish, it's corresponding mask showcasing BVG+ AND BVG- colonies and a zoom on touching colonies with boundary pixels in between. With thanks to T. Beznik.}
    \label{fig:colonies}
\end{figure}

We split the GSK dataset of Petri dish images gathered by \cite{Beznik2019} into train (65\%), validation (15\%) and test (20\%) sets. Each example is made of an image and its corresponding mask, as shown in Figure \ref{fig:colonies}.

Mask~R-CNN is an instance segmentation algorithm that combines a slightly updated version of Faster~R-CNN object detection algorithm \cite{ren2015faster} with convolution layers to further classify the pixels inside the detected bounding boxes into object and background pixels. Several key parameters were adapted to fit the nature of our use case. One of those is the lengths of square anchor sides. Indeed, in the RPN (Region Proposal Network), generating bounding boxes proposals adapted to the sizes of the objects we seek to predict is key to achieve a low loss. For instance, BVG objects are small compared to the Petri dish images and compared to other object recognition tasks.

Training was performed iteratively as follows\footnote{Training was achieved on a single machine equipped with an NVIDIA Tesla V100 GPU.}:

\begin{enumerate}
    \item \emph{epochs 0-399}: COCO pre-trained ResNet-50, original images, training network head layers only, 64 images per epoch
    \item \emph{epochs 400-449}: ResNet-50, original images, training all layers, 64 images per epoch
    \item \emph{epochs 450-499}: ResNet-50, augmented images, training all layers, 500 images per epoch
    \item \emph{epochs 500-554}: COCO pre-trained ResNet-101, augmented images, training all layers, 500 images per epoch
    \item \emph{epochs 555-600}: ResNet-101, strongly augmented images, training all layers, 500 images per epoch
\end{enumerate}

We first use a small, pre-trained backbone (ResNet-50) to speed-up the first training iterations, where only network head layers are trained since the backbone has already learned to recognise useful features on the COCO dataset \cite{lin2014microsoft}. We then unfreeze and train the backbone layers, which leads to good results on training data but mediocre performance on some validation images, likely due to the locations of CFUs on these images having never been seen in the training set. We then augment the training set in various ways: stochastic rotation of images from -180 to 180
~\degree, scalings, translations, and additive and multiplicative noises. We also increase the number of (now randomly generated) images to 500 per epoch which gives a good trade-off between training speed and periodic backups of the model's parameters. This leads to good model performance on the validation set. We then trade our ResNet-50 backbone for a 101 coco pre-trained one. A loss of performance is observed at first, as expected, but the model regains its previous performance level after only a few epochs and results slightly improve throughout the remaining epochs. We run a final training with more aggressive augmentations to make our model able to generalise to images that are very noisy, low contrast and taken from various angles and heights. This slightly improves the model's results even further up to a plateau on the validation set.

\section{Results}
\subsection{Best model without post-processing}

The left-hand block of Table \ref{table:performance} summarises our best model's results using the average mAP (mean Average Precision) on IoU (Intersection over Union) thresholds from 50 to 95\%, the mAP at IoU 50 and 70\%\footnote{To benchmark the overall performance of instance segmentation models, as defined and used in \cite{lin2014microsoft}.} and MAPE (Mean Absolute Percentage Error) on BVG-, BVG+ and all BVG counts\footnote{To assess the quality of model on the actual counting use case (the lower the better).}. One can see that the mAP at a 0.5~IoU threshold is close to 100\% (94.1\%) on the test set. Also, the overall MAPE (Mean Absolute Percentage Error) on the test set demonstrates that on average, the total counts on a given image is off by less than 3\%.

\begin{table}[H]
\caption{Benchmarks (in percentage) of our best model without post-processing (left) and with post-processing (right). The higher the mAP and the lower the MAPE, the better.}
\centering
\label{table:performance}
\begin{tabular}{llll|lll}
\toprule
{} & \multicolumn{3}{l|}{\textbf{No post-processing}} & \multicolumn{3}{l}{\textbf{Post-processing}}\\
{} & train &  val & test & train &  val & test \\
\midrule
mAP IoU=.50:.05:.95 & 58.3 & 51.9 & 50.7 & 58.2 & 51.0 & 50.6 \\
mAP IoU=.5          & 97.1 & 97.6 & 94.1 & 96.8 & 96.3 & 93.8 \\
mAP IoU=.75         & 63.5 & 47.5 & 50.5 & 63.2 & 46.8 & 50.6 \\
MAPE BVG-           & 7.7 & 2.2 & 12.5 & 9.3 & 14.4 & 14.3 \\
MAPE BVG+           & 3.1 & 2.1 & 5.7 & 2.8 & 1.9 & 4.8 \\
MAPE Tot            & 2.0 & 1.4 & 2.6 & 1.6 & 2.5 & 2.3 \\
\bottomrule
\end{tabular}
\end{table}

\subsection{Best model with post-processing}

We added several post-processing steps in order to further refine the quality of the results. First, we observed that the model is sometimes unsure about whether an object is a BVG+ or BVG-, and will typically generate two bounding boxes and masks for both classes. In those cases, we removed the least likely object, and marked the most likely one as unsure (BVG+ or BVG-) so that users can be easily notified about that and validate or invalidate our guess.

Second, CFUs on the border of Petri dishes are typically discarded by counting procedure rules. We implemented this feature by identifying the Petri boundary (i.e., the corresponding ellipse), shrinking it slightly and excluding predictions that do not intersect or are not included in this shrunken ellipse.

Third, in rare cases, dust grains on a Petri dish can be confused with CFUs. As dust grains are typically smaller than the colonies, we excluded predicted objects whose area can be considered as an outlier based on Laplace distributions percentiles computed for each image.

We searched the space of these post-processing features parameters in order to optimise model's performance on train and validation sets. We particularly focused on the total and BVG+ MAPEs as these are of utmost importance for the end-users.

The right-hand block of Table \ref{table:performance} shows our best model's results with the following post-processing steps applied to exclude the predicted BVGs in the following cases:

\begin{enumerate}
    \item segmented instance probability below 70\%
    \item least likely of two objects of different classes overlapping significantly (i.e., $IoU \geqslant~70\%$)
    \item outside of the 98\% version of the estimated ellipse delimiting the Petri dish
    \item outside of the 99\% confidence interval ([0.5\%, 99.5\%]) of the Laplace distribution fitted on the surface areas of BVGs on the current Petri dish image
\end{enumerate}

Compared to our model without post-processing (see left-hand side of Table \ref{table:performance}, we can see performance gains mainly on the MAPE, decreasing from 2.6\% to 2.3\% and from 5.7\% to 4.8\% for the total and BVG+ counts on the test set, respectively. This is achieved at the minor cost of a slight decrease in mAP performance (94.1\% to 93.8\%), again on the test set. 

Tables \ref{table:confusion-matrix-test-post-processing} and \ref{table:confusion-matrix-test-normalized-post-processing} show the original and normalised confusion matrices of the model's predictions with post-processing steps on the test set, respectively. As instance segmentation algorithms can predict non-existing objects or miss existing ones, an additional empty class is added. The intersection of these additional row and column (i.e., the last matrix element) always remains empty as non-predicted non-existing objects never occur. Furthermore, conversely to non-diagonal elements of standard classes, the \emph{Nothingness} elements, i.e., \emph{Missed} or \emph{Invented} objects, should be as low as possible as one never seeks to miss existing or invent non-existing objects.

With this in mind, one can see in Table \ref{table:confusion-matrix-test-normalized-post-processing} very few colonies get misclassified with less than 2\% and 7\% of actual BVG+ and BVG-, respectively. Also, one can note that actual BVG+ are less often confused with BVG- (0.5\%) than the converse (4.8\%).  Moreover, the model misses or invents twice as many BVG+ than BVG-, i.e., 3 to 6 and 2 to 4 colonies (see Table \ref{table:confusion-matrix-test-post-processing}). As the proportion of BVG+ colonies is more than twice that of the BVG- colonies, this shows that the model actually invents or misses fewer BVG+ colonies, relatively.

\begin{table}  
\centering
\caption{Confusion matrix on test set}
\label{table:confusion-matrix-test-post-processing}
\begin{tabular}{lrrrr}
\toprule
{} &  BVG- predicted &  BVG+ predicted &  Missed & Total \\
\midrule
BVG- actual        &              78 &               4 &                      2 &  84 \\
BVG+ actual        &               2 &             395 &                      4 &  401 \\
Invented &               3 &               6 &                    . &  9 \\
\bottomrule
\end{tabular}
\end{table}

\begin{table}
\centering
\caption{Normalised confusion matrix on test set (in \% of the actual number of CFUs)}
\label{table:confusion-matrix-test-normalized-post-processing}
\begin{tabular}{lrrrr}
\toprule
{} &  BVG- predicted &  BVG+ predicted &  Missed & Total \\
\midrule
BVG- actual        &            92.9 &             4.8 &                    2.4 &  100 \\
BVG+ actual        &             0.5 &            98.5 &                    1.0 &  100 \\
Invented &            33.3 &            66.7 &                    .  &  100 \\
\bottomrule
\end{tabular}

\end{table}

\subsection{Visual analysis of the our results}

\begin{figure}
    \centering
    \includegraphics[width=0.9\textwidth]{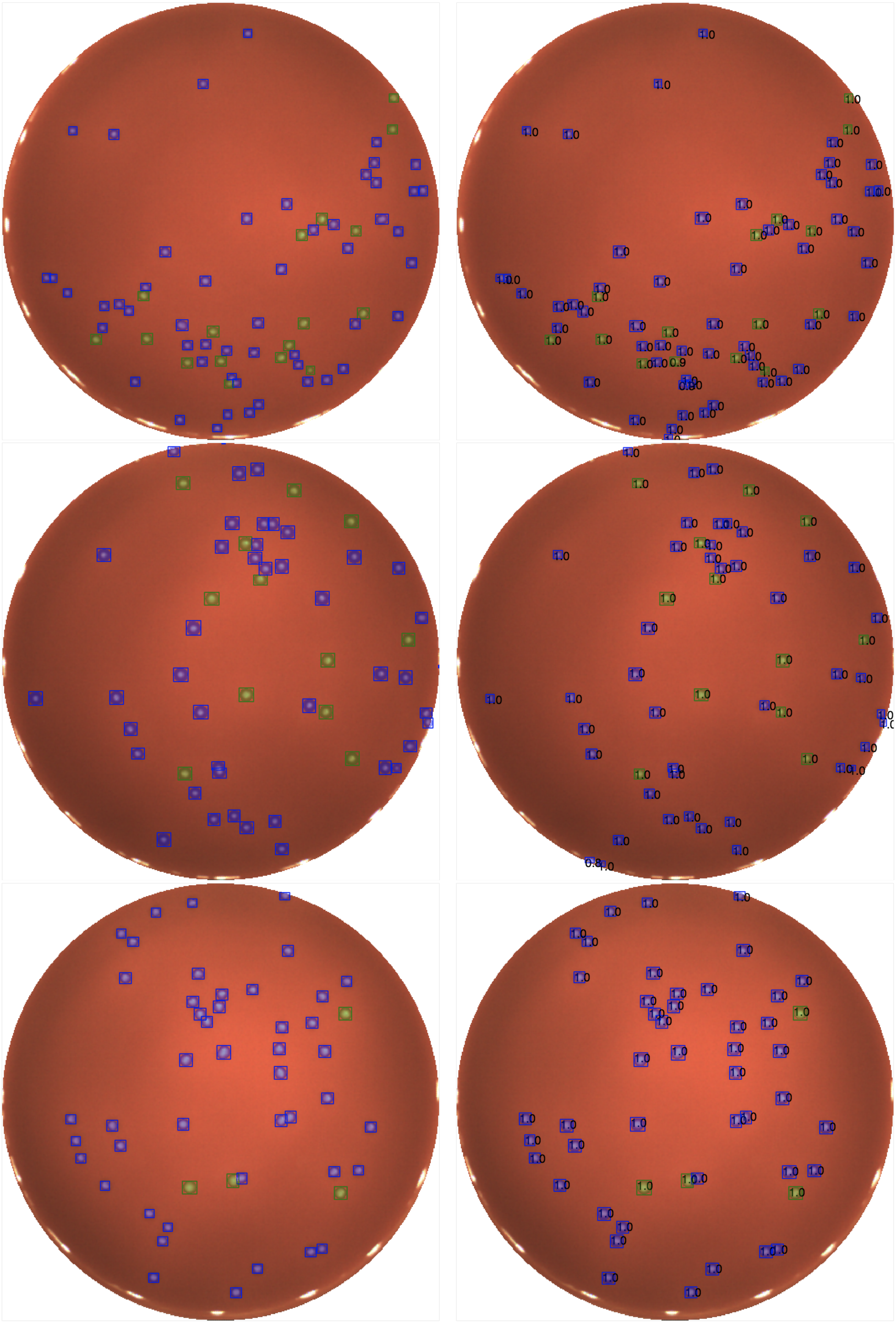}
    \caption{Model predictions (left) and exported actual BVG quantities (right) in our app interface. Test data (on the left) and our corresponding best model's predictions and probabilities (on the right)}
    \label{fig:predictions-test-post-processing}
\end{figure}

Figure \ref{fig:predictions-test-post-processing} shows some examples of predicted CFUs on our test set with our best model and we can see that the results are very promising. Most CFUs are identified by the model and correctly classified.

Further inspection shows that most of the errors arise from CFUs on the images' borders. One of our post-processing steps identifies the Petri dish border and only includes predictions within a 98\% reduced ellipse of the border. The reason to not further reduce the ellipse to dismiss predictions on the border lies in the fact that the images taken by the camera are slightly asymmetrically cropped. This results in some images not completely encompassing some borders of the Petri dish. For instance, this is particularly noticeable on the bottom and left borders of the two first images presented in Figure \ref{fig:predictions-test-post-processing}. Hence, using a greatly reduced ellipse of the border would lead to missing legitimate BVGs on the top and right borders. As an intermediate solution, we display a conservative boundary ellipse and let the users move, expand or reduce it to better accommodate the asymmetric cropping of the images. In the long run, tablets could eventually be used instead of the laboratory camera, and images that do not crop the borders of the Petri dish should be easily gathered, solving this problem by itself.

\subsection{User-to-User and User-to-Model variability.}

We conducted an experiment to compare the agreement among users and between users and our model. To do so, we asked two lab technicians to independently label a newly cultivated batch of 36 Petri dish images and also ran our model on those images. In turn, this allowed us to compare the variability between the users and between the users and the model for the total number of CFUs, as well as the BVG+ and BVG- ones. Those results are displayed on Table \ref{table:users_vs_model} and show that users disagree less with our model than among themselves for BVG+ and BVG- counts, and to a similar extend for the total counts.

\begin{table}[H]
\caption{User-to-User and User-to-Model variability.}
\centering
\label{table:users_vs_model}
\begin{tabular}{lrrrr}
\toprule
{} &  User to User variability &  Users to Model variability \\
\midrule
MAPE Total count        &              16 &  16 \\
MAPE BVG+ count        &              43 &  33 \\
MAPE BVG- count        &              68 &  45 \\
\bottomrule
\end{tabular}
\end{table}

Our results show that the users disagree less with the model than with each other. 

\section{Single-Page-Application}
\label{web-application}

Our web-application is developed with the Vue.js framework \cite{vuejs} and a Vuetify component-based layout \cite{vuetify}. The first iteration focused on a minimal interface allowing users to load images and examine them in an interface allowing zooming, laying out BVG- and BVG+ colonies. A first Deeplabv3 semantic segmentation model was quickly trained on the data and further post-processed to roughly identify the colonies so that the front-end could trigger inferences for newly updated images and present those raw first results. The second iteration improved upon the interface to examine images. Users were then able to also modify, delete and create CFUs and the modifications got persisted. The third iteration allowed users to specify dilution factors for each (or groups of) images, generate and export the actual number of bacteria and their confidence interval\footnote{End-users experiments consistently rely on batches of 3 triplicates of Petri dishes, each triplicate associated with a different dilution factor to estimate the total count of bacteria and its confidence interval.}.

\begin{figure}[h]
    \centering
    \includegraphics[width=\textwidth]{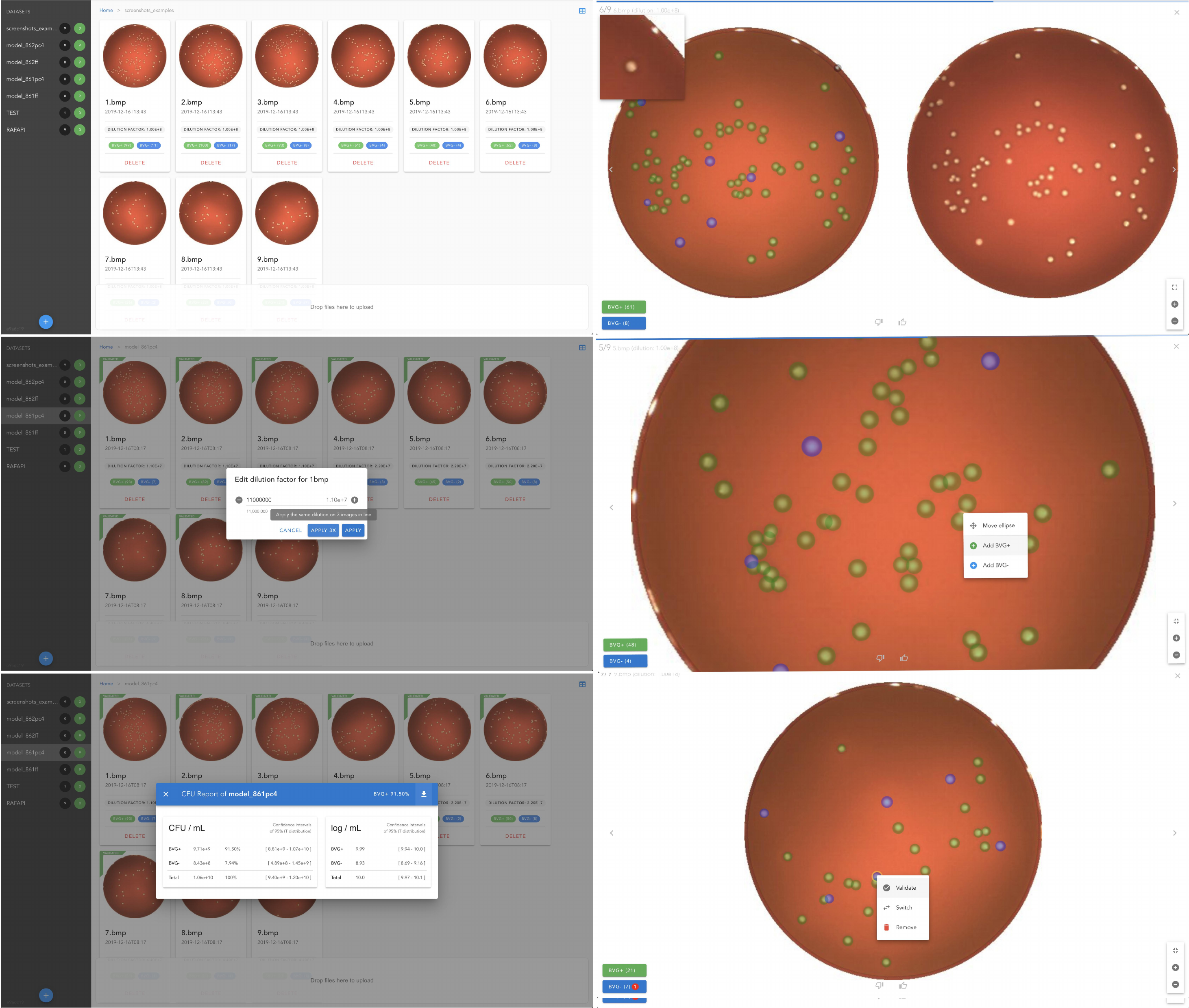}
    \caption{Screenshots of the Single-Page-Application. Left sub-figures depict, from top to bottom, a dataset overview, modifications to dilution factors and exported actual BVG quantities. Right sub-figures show, from top to bottom, an overview of a Petri Dish with its predictions, an overview with deselected BVG+ with zoomed hoovering and a zoomed overview with new BVG editing.}
    \label{fig:web-app-screenshots}
\end{figure}

Uncertain CFUs were shown to users with the predicted class so they could easily validate, invalidate, or remove a prediction. Warning and error messages were displayed to the user if inconsistent dilution factors\footnote{E.g., not decreasing dilution triplicates.} or number of images were created within an experiment. Finally, we displayed the border of the Petri dishes and allowed users to change it if deemed necessary, to include or exclude more colonies on the borders. Figure \ref{fig:web-app-screenshots} shows a selection of screenshots of the web application.

\section{Conclusions and future work}

We have shown how a Mask~R-CNN model can be tuned and trained to successfully count BVG- and BVG+ bacterial CFUs. Our approach solves the imbalanced class problem depicted in Section \ref{Introduction}, and significantly alleviates the problem caused by overlapping CFUs. Additional post-processing steps can also be used to further improve the quality of the predictions. Data augmentation, mainly rotations, are key to getting good performance with very few training examples. More drastic transformations help the model better perform on poor exposure conditions, as well as on pictures taken at different angles and heights. 

We also successfully developed a single-page-application that lab technicians can use to reliably speed-up their workflow. Users can then further validate and modify results as they deem necessary. Finally, dilution factors can be associated to Petri dish images to further infer the quantities of BVG- and BVG+ bacteria and their confidence intervals.

Further improvements in the predictions quality could be achieved with more labelled data, although current results are already satisfactory to the operators. Moreover, the methodology and the app could be extended to more species of CFUs as well as viral foci. Another add-on could be setting-up a user-friendly interface embedded in our app so users can annotate new datasets. This would in turn facilitate the training and generalisation to other species of bacteria. Finally, the areas of colonies are already computed and could later on be used to better estimate the actual quantity of bacteria in dishes.

\section{Disclosure}

This work was sponsored by GlaxoSmithKline Biologicals SA. All authors were involved in drafting the manuscript and approved the final version. The authors declare the following interest: GDL and PS are employees of the GSK group of companies and report ownership of GSK shares and/or restricted GSK shares. TN, LS, MH are employees of Radix.ai, a consultancy firm contracted by GlaxoSmithKline Biologicals SA in the context of this study.  The authors would like to thank Sebastien Cheffert, Thomas  Haupt, Amin Khan, Sabine Leclercq,  Alex Pysik and Gurpreet Singh for their valuable input.

\bibliographystyle{unsrt}  


\end{document}